\documentclass[a4paper,conference]{IEEEtran}
\usepackage{cite}

\usepackage{amsthm, amssymb, amsmath}
\usepackage{graphicx}
\usepackage{float}
\usepackage{rotating}
\usepackage{subfigure}

\usepackage[justification=centering]{caption}
\usepackage[lined,ruled,linesnumbered,commentsnumbered]{algorithm2e}
\usepackage{cleveref}
\usepackage{booktabs}

\usepackage{pdflscape}

\usepackage{color, colortbl}

\newtheorem{definition}{Definition}

\newtheorem{remark}{Remark}

\usepackage{amsmath}
\usepackage{array}
\usepackage{url}

\IEEEoverridecommandlockouts

\begin{document}

\bibliographystyle{IEEEtran}
%

\title{\ \\ \LARGE\bf Solving Dynamic Multi-objective Optimization Problems via Support Vector Machine  \thanks{Min JIANG, Weizhen HU, Liming QIU, and Minghui SHI are with the Department of Cognitive Science and Fujian Key Laboratory of Brain-inspired Computing Technique and Applications (Xiamen University). }
	\thanks{Kay Chen Tan is with the Department of Computer Science of City University of Hong Kong.  }
	\thanks{The Corresponding author: Dr. Minghui SHI, smh@xmu.edu.cn.}
}

\author{
\IEEEauthorblockN{1\textsuperscript{st} Min JIANG}
\IEEEauthorblockA{Department of Cognitive Science and Technology\\
Xiamen University\\
Xiamen, China \\
minjiang@xmu.edu.cn}
\and
\IEEEauthorblockN{2\textsuperscript{nd} Weizhen HU}
\IEEEauthorblockA{Department of Cognitive\\Science and Technology\\
Xiamen University\\
Xiamen, China \\
hwz@stu.xmu.edu.cn}
\and
\IEEEauthorblockN{3\textsuperscript{rd} Liming QIU}
\IEEEauthorblockA{Department of Cognitive Science and Technology \\
Xiamen University\\
Xiamen, China \\
qlm@stu.xmu.edu.cn}
\and
\IEEEauthorblockN{4\textsuperscript{th} Minghui SHI}
\IEEEauthorblockA{Department of Cognitive Science and Technology \\
Xiamen University\\
Xiamen, China \\
smh@xmu.edu.cn}
\and
  \IEEEauthorblockN{5\textsuperscript{th} Kay Chen TAN}
\IEEEauthorblockA{Department of Computer Science \\
City University of Hong Kong\\
Hong Kong, China \\
kaytan@cityu.edu.hk}

}
\IEEEoverridecommandlockouts
\IEEEpubid{\makebox[\columnwidth]{978-1-5386-4362-4/18/\$31.00~\copyright2018
IEEE \hfill} \hspace{\columnsep}\makebox[\columnwidth]{ }}
\maketitle
\begin{abstract}
Dynamic Multi-objective Optimization Problems (DMOPs) refer to optimization problems that objective functions will change with time. Solving DMOPs implies that the Pareto Optimal Set (POS) at different moments can be accurately found, and this is a very difficult job due to the dynamics of the optimization problems. The POS that have been obtained in the past can help us to find the POS of the next time more quickly and accurately. Therefore, in this paper we propose a Support Vector Machine (SVM) based Dynamic Multi-Objective Evolutionary optimization Algorithm, called SVM-DMOEA. The algorithm uses the POS that has been obtained to train a SVM and then take the trained SVM to classify the solutions of the dynamic optimization problem at the next moment, and thus it is able to generate an initial population which consists of different individuals recognized by the trained SVM. The initial populuation can be fed into any population based optimization algorithm, e.g., the Nondominated Sorting Genetic Algorithm II (NSGA-II), to get the POS at that moment. The experimental results show the validity of our proposed approach.
\end{abstract}


%
\IEEEpeerreviewmaketitle

\section{Introduction}
The objective functions of Dynamic Multiobjective Optimization Problems (DMOPs) \cite{Farina_2004}  will change with time, and this characteristic bears significant implications for lots of real-world applications \cite{Cruz_2010}, so how to solve the DMOPs has attracted a great deal of attention from researchers in related fields. However, the existing approaches often do not work well in solving this problem since that the Pareto Optimal Set (POS) of a DMOP keeps changing. Solving the DMOPs efficiently and effectively has become an important research topic in evolutionary computation community \cite{Nguyen_2012}.

Generally, once the population converges, it is difficult to adapt to changes in the environment, and this is one of the reasons why evolutionary algorithms do not perform well in handling dynamic optimization problems. How to effectively reuse "experiences" has become the way to solve this problem. For example, The prediction based method proposed in \cite{Rossi_2008} predicts the changing optimization functions by using some machine learning techniques, and the basic idea of  is ``keeping track of good (partial) solutions in order to reuse them under periodically changing environment''.  Zhou \textit{et al.} \cite{Aimin_Zhou_2014} presented an algorithm, called Population Prediction Strategy (PPS), to predict a whole population instead of predicting some isolated points.

Our interesting point is placed on how EA can quickly re-optimize a given dynamic optimization problem once the change is been detected and identified. At the same time, some recent research results \cite{Dang_2015,dang2016populations} show that the population plays a very important role for tracking dynamic optima, so in this research, we study how to use a machine learning technique, says Support Vector Machine (SVM), to generate an initial population, which can help us solving dynamic multi-objective optimization problems.

The contribution of this research is the integration between a proven machine learning technique and classical evolutionary multi-objective optimization algorithms. This combination provides two benefits. First, this approach can preserve the advantages of the EAs and the SVM; Secondly, the proposed design can improve the search efficiency by reusing past experience which is critical for solving the DMOPs.

The rest of this paper is organized as follows: In Section \ref{sec:Preliminaries-and-Related}, we will introduce some basic concepts of dynamic optimization problems and the support vector machine . In Section \ref{sec:SVM-DMOEA}, we will propose the Support Vector Machine based Dynamic Multi-Objective Evolutionary optimization Algorithm, SVM-DMOEA. In Section \ref{experiments} we firstly describe evaluation standards, test instances and comparative methods, and the latter part of Section \ref{experiments}, the experimental results and some discussions are presented. In the Section \ref{Conclusion}, we conclude the main work of this research and future research direction is also discussed.

\section{Preliminaries and Related Works}
\label{sec:Preliminaries-and-Related}

\subsection{Concepts of Multi-objective Optimization}
In this research, we just consider dynamic multi-objective optimization problems, and the optimization problem is defined as:
$$\textbf{Minimize}~F\left(x,t\right)=\left<f_{1}\left(x,t\right),f_{2}\left(x,t\right),...,f_{M}\left(x,t\right)\right>$$
$$s.t.\ x\in\Omega, $$
where $x=\left<x_{1},x_{2},\ \ldots,\ x_{n}\right>$ is the decision vector and $t$ is the time or environment variable. $f_{i}\left(x,t\right):\Omega\ \rightarrow\ \mathbb{R}\ \left(i=1,\ \ldots,\ M\right)$.  $\Omega = \left[L_1,U_1\right] \times \left[L_2,U_2\right] \times \cdots \times \left[L_n,U_n\right]$. $L_i, U_i \in \mathbb{R}$ are the lower and upper bounds of the $i$-th decision variable, respectively.

\begin{definition}{\emph{[Dynamic Decision Vector Domination]}}
	At time $t$ , a decision vector $x_{1}$ Pareto dominate another vector $x_{2}$, denoted by $x_{1}\succ_{t}x_{2}$, if and only if\emph{:}
	\begin{equation}
	\begin{cases}
	\forall i=1,\ldots,M, & f_{i}(x_{1},t)\leq f_{i}(x_{2},t)\\
	\exists i=1,\ldots,M, & f_{i}(x_{1},t)<f_{i}(x_{2},t)
	\end{cases}.
	\end{equation}
\end{definition}

\begin{definition}{\emph{[Dynamic Pareto-optimal Set]} }
	Both $x$ and $\ensuremath{x^{*}}$ are decision vectors, and a decision vector $x^*$ is said to be nondominated at time $t$ if and only if there is no other decision vector $x$ such that $x \succ_t x^*$ at time $t$. The Dynamic Pareto-Optimal Set (DPOS) is the set of all Pareto optimal solutions at time $t$, that is\emph{:}
	$$DPOS=\left\{ x^{*}|\not \exists x,~x\succ_{t}x^{*}\right\}.$$
\end{definition}

\begin{definition}{\emph{[Dynamic Pareto-optimal Front]}}
	At time $t$, the Dynamic Pareto-Optimal Front (DPOF) is the corresponding objective vectors of the DPOS.	
	$$DPOF=\left\{ F\left(x^{*},t\right)|x^{*}\in DPOS\right\}.$$
\end{definition}

\subsection{Support Vector Machines}
Support Vector Machine (SVM) was first introduced in 1992 \cite{boser1992training}, and it is related to statistical learning theory \cite{vapnik2013nature}. SVM becomes a well-known learning method used for classification problems. The basic idea of the original one is to find a hyperplane which separates the $d$-dimensional data perfectly into its two classes.

In SVM, training is reformulated to obtain a quadratic programming (QP) problem. Let $\{x_1, \cdots, x_t\}$ be training data set and $y_i \in \{1,-1\}$ be the class label of $x_i, i = 1, \cdots, t$. The data are mapped by a function $\phi: \mathbb{R}^t \rightarrow F$, called feature mapping, into a ``feature space'', and the linear hyperplanes that divide the data into two labeled classes can be shown as:
$$\mathbf{w^T} \times \phi(x)+b=0, $$ where $\mathbf{w}\in \mathbb{R}^t$ and $b\in \mathbb{R}$.

To get an optimal hyperplane with maximum-margin and bounded error in the training data, we can solve the following QP problem:
\begin{eqnarray}
&\mathbf{Minmize_{(w,b)}}~\frac{1}{2}||\mathbf{w}||^2 + C\cdot \sum_{i=1}^{t}\varepsilon_i \nonumber\\
& \mathbf{ subj.~to:~} y_i(\mathbf{w}\times\phi(x_i)+b)\geq 1-\varepsilon_i, i = i \cdots t.
\label{SVM}
\end{eqnarray}
Where $C$ is a constant and the second term of Equation (\ref{SVM}) provides an upper bound for the error in the training data, and the first term  makes maximum margin of separation between classes.

We can solve Equation (\ref{SVM}) by using Lagrange method. Given a kernel function $\kappa(x_i,x_j)=\phi(x_i)\cdot \phi(x_j)$, the Lagrange function of Equation (\ref{SVM}) is as follows:
\begin{eqnarray}
&\mathbf{maximize_{(\alpha)}}~ \sum_{i=1}^{t}\alpha_i -\frac{1}{2}\sum_{i,j=1}^{t}\alpha_i\alpha_jy_iy_j\kappa(x_i,x_j) \nonumber\\
& \mathbf{ subj.~to:~} \mathbf{w} = \sum_{i=1}^{t}y_i\alpha_i\phi(x_i), \sum_{i=1}^{t}\alpha_iy_i=0 \nonumber\\
&0\leq \alpha_i\leq C
\label{SVM-1}
\end{eqnarray}

Equation (\ref{SVM-1}) shows that the optimal hyperplane can be regarded as the linear combination of training samples with $\alpha_i \neq 0$. These samples are called support vectors and build the decision function of the classifier based on the kernel function

\begin{equation}
f(x)= \mathbf{sgn}\Big(\sum_{i=1}^{t}y_i\alpha_i\kappa(x,x_i)\Big).
\end{equation}
There are lots of different kernel functions, and they are often selected based on the features of testing data and type of the boundaries between classes. The most popular ones are linear kernel function, polynomial kernel functionn and Gaussian kernel function. More details about the SVM are available in \cite{suykens1999least}.

\subsection{Related Works}
Great progress \cite{Azzouz_2016,azzouz2017dynamic} has been made in the DMOPs field, and these algorithms can be classified into the following categories roughly: Increasing/Maintaining Diversity methods, Memory based methods, Multi-population based methods, and Prediction based methods.

The increasing diversity methods tend to add variety to the population by using a certain type of methodology when the environment change was detected. For example, Cobb \textit{et al.} proposed the triggered hypermutation method \cite{cobb1990investigation}, and the basic idea of this method is that when change is identified, the mutation rate would be increased immediately, and this would make the converged population divergent again. This approach calls for some improvements, and one of them is that the mutation rate is in a state of uncontrolled change during the whole process, and this ultimately results in reduced performance of the algorithm. Therefore, Yen \textit{et al.} \cite{Woldesenbet_2009} proposed a dynamic EA which relocates the individuals based on their change in function value due to the change in the environment and the average sensitivities of their decision variables to the corresponding change in the objective space. This approach can avoid the drawbacks of previous methods to a certain extent.

Dynamic NSGA-II (DNSGA-II) \cite{Deb}  proposed by Deb \textit{et al.} also shares a similar idea, and this method handles the DMOPs by introducing diversity when change is detected. There are two versions of the proposed DNSGA-II and they are respectively known as DNSGA-II-A and DNSGA-II-B. In  the DNSGA-II-A, the population is replaced by some individuals with new randomly created solutions, while in the DNSGA-II-B, diversity was guarded by replacing a percentage of the population with mutated solutions.

Memory mechanism enables EAs to record past information, and when it detects changes have occurred, stored information can be reused to improve the performance of the algorithm. Existing research showed that memory-based approaches tend to be more effective on the DMOPs with periodically changing environments. Branke \cite{Branke} proposed a direct memory scheme where the best individuals in the population will be saved in an archive, and when the algorithm detects a change, those saved individuals can be retrieved and returned to the population to replace the same number of individuals. In \cite{Azzouz_2015}, the authros proposed an adaptive hybrid population management strategy using memory, local search and random strategies to effectively handle dynamicity in DMOPs.

The Multi-population strategy is considered as one efficient solution for the DMOPs, especially for the multiple peaks and the competing peaks problems. Li and Yang \cite{Li_2008} employed a multi-population particle swarm optimization (PSO) algorithm to solve multiple peaks problems. In their method, a population uses evolutionary programming, which shows a better global search ability when compared to other EAs, to explore the most hopeful areas in the whole search space.

In recent years the prediction-based DMOPs algorithms have received much attention. Bosman \cite{Bosman_2007} believed that the decision made at one point would affect the optima obtained in the future, so for the dynamic optimization problems, he proposed an algorithmical framework which integrated machine learning, statistic learning, and evolutionary computation, and this framework can effectively predict what the state of environment is going to be. In \cite{Aimin_Zhou_2014}, Zhou \textit{et al.} presented an algorithm, called Population Prediction Strategy (PPS), to predict a whole population instead of predicting some isolated points.

In \cite{Muruganantham_2016}, the authors proposed a dynamic multiobjective evolutionary algorithm, MOEA\textbackslash{}D-KF, which used Kalman filter (KF) to predict the moving optima  in decision space. A novelty of the approach is that a scoring scheme is designed to hybridize the KF prediction with a random reinitialization method. The predictions help to guide the search toward the changed optima, thereby accelerating convergence.

\section{Solving Dynamic Multi-objective Optimization Problems via Imbalanced Data Learning Method And Incremental Support Vector Machine}
In this section, we propose a SVM based dynamic optimization algorithm. Our motivation is that to get a solution of a dynamic optimization problem is an easy task, however judging whether this solution is a ``good'' solution is a very hard thing. So we convert this decision problem into a classification problem. In other words, we use the knowledge we have gained, the solutions in the POS in the past, to train an SVM classifier and then use this classifier to classify the randomly generated solutions for the dynamic optimization problem in the next time, thereby we can obtain an initial population, and this population can help us to improve the efficiency of solving dynamic multi-objective optimization problems.

under different environments obey  different  probability distributions, and these distributions are not identical but are correlated.
In this section, we will explain the proposed algorithm, SVM-DMOEA, in detail.

In the first place, we introduce the basic idea of this algorithm. A good population is crucial for solving dynamic multi-objective optimization problems. A good population not only speeds up the solution, but also improves the quality of the solution. The experience gained from obtaining the POS in the past can help us to find an initial population. It is easy task to find a solution to a dynamic multi-objective problem, however, it is a hard problem how to judge whether the solution we found is ``good’’ enough. If a solution is ``good’’, we can put them into an initial population which can be used to compute the POS of next time.

We consider the problem of judging whether a solution can become an individual in the initial population as a classification problem, such that we can train an SVM by using solutions in the POS just obtained (positive examples) and the ones that are not POS (negative examples) . Using this SVM, we can classify the solution to the next moment which are generated at random into two classes – “good” or “not that good”. The solutions that have passed the recognition of the SVM can be put into the initial population of the next moment. This initial population can be fed into any population-based evolutionary algorithm to get the POS at the next time.

\label{sec:SVM-DMOEA}
\begin{algorithm}
	\caption{SVM-DMOEA: Support Vector Machines based Dynamic Multi-objective Evolutionary Algorithm}
	\label{alg:SVM-DMOEA}
	\KwIn{ The Dynamic Multi-objective Optimaztion Function $F(X)$;  }
	\KwOut{$POSs$: the POSs of $F(X)$;}
	Randomly initiate a Population the $Pop_0$\;
	$POS_0$ = NSGA-II($Pop_0$)\;
	$POS_s$	=$POS_0$\;
	\For{$t = 0$ to $n$}{
		Train a SVM classifier $SC$ by using $g \in POS_{t}$ and $ng \notin POS_{t}$\;
		Randomly generate solutions $\{xy_1,\cdots,xy_{p}\}$ of the function $F(X)_{t+1}$\;
		\If{$xy_i$ pass the recognition of the SVM $SC$ }{Put $xy_i$ into $Pop_{t+1}$}
		$POS_{t+1}$ = NSGA-II($Pop_{t+1}$)\;
		$POSs$ = $POSs \cup POS_{t+1}$ \;		
	}
	\Return $POSs$\;
\end{algorithm}

\begin{remark}
	Please note that in Algorithm \ref{alg:SVM-DMOEA}, we use NSGA-II algorithm to obtain the POS based on the initial popluation we found, however in reality, NSGA-II can be replaced by any population-based multi-objective algorithm.
\end{remark}

\section{Empirical Study}
\label{experiments}
\subsection{Performance Metrics, Testing Functions and Settings}
In this research, we use a variant of the Inverted Generational Distance (IGD) to evaluate  the quality of the solutions obtained by these competing algorithms.

The inverted generational distance (IGD) \cite{sierra2005improving} is a metric  to quantify the performance of a multiobjective optimization algorithm. Let $P^*$ be the set of uniformly distributed Pareto optimal solutions in the POF and $P$ represent the POF obtained by the algorithm,  the definition of the IGD is：
	\begin{equation}
	\label{def:igd}
	\mathrm{IGD}(P^*, P, C) = \frac{\sum_{v^* \in P^*}\min_{v\in P}\left\|v^*-v\right\|}{\left|P^*\right|}.
	\end{equation}
	
If the $P$ is close enough to $P^*$, the value of IGD will be as small as possible. In other words, the IGD depicts the difference between the ideal POF and the POF obtained by the competing algorithms.
	
	Please note that the definition of the IGD is slightly different from the original one, and the major difference is the parameter $C$ in Equation (\ref{def:igd}). The parameter $C$ is a combination of the benchmark functions parameters. We call it as configuration of the benchmark functions. The configurations we used in our experiments are described in Table \ref{table:settings}.
	
 One variant of the IGD, called MIGD, can also be used to evaluate dynamic multiobjective optimization algorithms \cite{Muruganantham_2015,Muruganantham_2016} , and it takes the average of the IGD values in some time steps over a run as the performance metric, given by
	\begin{equation}
	\mathrm{MIGD}(P^*, P, C) = \frac{1}{|T|}\sum_{t \in T}\mathrm{IGD}(P_t^{*}, P_{t}, C),
	\end{equation}
	where $P_t^{*}$ and $P^{t}$ represent the points set of the ideal POF and the approximate POF obtained by the algorithm at time $t$. We also want to evaluate those algorithms under different environments, so a novel metric, DMIGD, is defined based on the MIGD, and the definition of the DMIGD is as follows:
	
	\begin{equation}
	\mathrm{DMIGD}(P^*, P, C) = \frac{1}{|E|}\sum_{C \in E}\mathrm{MIGD}(P_t^{*}, P_{t}, C),
	\end{equation}
	where $|E|$ is the number of the different environments experienced. In our experiments, we choose eight different configurations. As a result, $|E|$ equals to eight. What we want to point out is that the DMIGD can evaluate a dynamic optimization algorithm from a  high-level view and it bears a significant difference with the MIGD since the MIGD just considers the dynamics in one environment.
	
\subsection{Test Instances and Experimental Settings}
In this research, we take the IEEE CEC 2015 benchmark problems set as test functions and the problem set has twelve testing functions. Details of the functions definitions are given in \cite{helbig2015benchmark}.  In the definitions, the decision variables are $x=(x_1,\ldots,x_n)$ and $t = \frac{1}{n_t}\left\lfloor \frac{\tau_T}{\tau_t} \right\rfloor$, where $n_t, \tau_T$, and $\tau_t$ are the severity of change, maximum number of iterations, and frequency of change respectively. Table \ref{table:settings} describes the different combinations of $n_t$, $\tau_t$, and $\tau_T$ used in our experiments. Please note that, for each $n_t$-$\tau_T$ combination, there will be $\frac{\tau_T}{\tau_t}$ environment changes. In other words, in all of our experiments, there are altogether five changes for the twelve dynamic problems.

\begin{table}[!htbp]
	\centering
	\caption{Environment Settings}
	\label{table:settings}
	\begin{tabular}{cccc}
		\toprule
		& \textbf{$n_t$} & \textbf{$T_t$} & \textbf{$T_T$} \\
		\midrule
		\textbf{C1} & 10    & 5     & 25 \\
		\textbf{C2} & 10    & 10    & 50 \\
		\textbf{C3} & 10    & 25    & 125 \\
		\textbf{C4} & 10    & 50    & 250 \\
		\textbf{C5} & 1     & 5    & 25 \\
		\textbf{C6} & 1     & 10    & 50 \\
		\textbf{C7} & 20    & 25    & 125 \\
		\textbf{C8} & 20    & 50    & 250 \\
		\bottomrule
	\end{tabular}%
\end{table}

The POFs of the  testing functions have different shapes and each function belongs to a certain DMOPs type. For example, the POF of the functions could be non-convex, convex, isolated, deceptive, continuous or discontinuous. Table \ref{table:instances} describes the types of the testing functions. Type I implies POS changes, but POF does not change; Type II means that the POS and the POF change as well; Type III means that the POF changes, but the POS does not change.

\begin{table}[!htbp]
	\caption{Characteristic of the test functions}
	\label{table:instances}
	\centering
	\begin{tabular}{|l|c|c|c|}
		\hline
		Name & \begin{tabular}{c}Decision \\Variable \\Dimension\end{tabular} & Objectives & DMOP Type \\\hline
		FDA4          & 12 & 3 & TYPE I   \\\hline
		FDA5          & 12 & 3 & TYPE II  \\\hline
		FDA5$_{iso}$  & 12 & 3 & TYPE II  \\\hline
		FDA5$_{dec}$  & 12 & 3 & TYPE II  \\\hline
		DIMP2         & 10 & 2 & TYPE I   \\\hline
		dMOP2         & 10 & 2 & TYPE II  \\\hline
		dMOP2$_{iso}$ & 10 & 2 & TYPE II  \\\hline
		dMOP2$_{dec}$ & 10 & 2 & TYPE II  \\\hline
		dMOP3         & 10 & 2 & TYPE I   \\\hline
		HE2                   & 30 & 2 & TYPE III \\\hline
		HE7                   & 10 & 2 & TYPE III \\\hline
		HE9                   & 10 & 2 & TYPE III \\\hline
	\end{tabular}
	
\end{table}

As depicted in Table \ref{table:instances}, the dimensions of the decision variables are from 10 to 30, and the parameters A, B and C for the functions FDA5$_{iso}$, FDA5$_{dec}$, dMOP2$_{iso}$ and dMOP2$_{dec}$ are set to G(t), 0.001 and 0.05 respectively.	

\subsection{Experimental Results}
In this research, we compare the SVM-DMOEA with six chosen algorithms.  A random reinitialization method (RND) \cite{Muruganantham_2016} is implemented as a baseline. Dynamic NSGA-II (DNSGA-II) \cite{deb2007dynamic} proposed by Deb \emph{et al.} handles the DMOPs by introducing diversity when change is detected. There are two versions of the proposed DNSGA-II and there are respectively known as DNSGA-II-A and DNSGA-II-B. In DNSGA-II-A, the population was replaced by some individuals with new randomly created solutions, while in the DNSGA-II-B diversity was guarded by replacing a percentage of the population with mutated solutions. The MBN-EDA \cite{karshenas2014multiobjective} is a multi-objective estimation of distribution algorithm and uses the multi-dimensional Bayesian network as its probabilistic model to capture the dependencies between decision variables and target variables. We also compare  the proposed algorithm with other dynamic multi-objective optimization algorithms. MOEA\textbackslash{}D-KF was presented in \cite{Muruganantham_2016},  and this method uses Kalman Filter to make predictions in decision space to solve DMOPs.

\begin{table*}[htbp]
	\centering
	\caption{Statistical results of the proposed algorithm, SVM-DMOEA, and six chosen competing algorithms in term of DMIGD performance metric}
	\label{tab:DMIGD}%
	\renewcommand\arraystretch{1.2}
	\resizebox{1\textwidth}{!}{
	\begin{tabular}{lccccccc}
		\toprule
		\toprule
		\multicolumn{1}{c}{\textbf{DMIGD}} & \multicolumn{1}{c}{\textbf{NSGA2}} & \multicolumn{1}{c}{\textbf{DNSGA-II-A}} & \multicolumn{1}{c}{\textbf{DNSGA-II-B}} & \multicolumn{1}{c}{\textbf{MBN-EDA}} & \multicolumn{1}{c}{\textbf{RND}} & \multicolumn{1}{c}{\textbf{MOEA\textbackslash{}D-KF}} & \multicolumn{1}{c}{\textbf{SVM-DMOEA}} \\
		\midrule
		FDA4  & 0.2634 & 0.3110 & 0.2790 & 0.4300 & 0.1698 & 0.1810 & \textbf{0.1660} \\
		FDA5  & 0.3301 & 0.3720 & 0.3710 & 0.5100 & 0.5323 & 0.3060 & \textbf{0.2910} \\
		FDA5\_iso & 0.1048 & \textbf{0.0840} & 0.0880 & 0.6400 & 0.1433 & 0.1000 & 0.2451 \\
		FDA5\_dec & 0.5923 & 0.5850 & 0.5430 & 1.2700 & 0.5403 & \textbf{0.2950} & 0.3630 \\
		DIMP2 & 3.8986 & 3.6910 & 4.3040 & 6.9700 & 17.9537 & 5.0690 & \textbf{2.9814} \\
		DMOP2 & 0.4202 & 1.2210 & 1.1790 & 1.4000 & 1.4329 & 1.4870 & \textbf{0.0695} \\
		DMOP2\_iso & 0.0325 & 0.0270 & \textbf{0.0260} & 2.5600 & 0.0315 & 0.0190 & 0.0455 \\
		DMOP2\_dec & 0.6303 & 0.7290 & 0.7820 & 2.8900 & 9.0504 & 3.4770 & \textbf{0.2040} \\
		DMOP3 & 0.8851 & 0.1870 & 1.0890 & 1.3800 & \textbf{0.0697} & 6.7890 & 0.0824 \\
		HE2   & 0.2096 & 0.2120 & 0.2160 & 0.8300 & \textbf{0.0744} & 0.1800 & 0.0910 \\
		HE7   & 0.0946 & 0.0820 & 0.0830 & 0.2100 & 0.1787 & 0.1510 & \textbf{0.0728} \\
		HE9   & 0.2945 & 0.2850 & 0.2840 & 0.3600 & 0.3432 & 0.3320 & \textbf{0.2497} \\
		\bottomrule
	\end{tabular}%
}
\end{table*}%
Table \ref{tab:DMIGD} records the DMIGD values of the algorithms running on different testing functions under different configurations. For the twelve test functions, the SVM-DMOEA achieves the best results in seven benchmark functions, says FDA4, FDA5, DIMP2, DMOP2, DMOP$_{dec}$, HE7 and HE9. The experiments seem to show that the proposed algorithm is competitive at handling the Types II and III DMOPs since major parts of these test functions belong in these two types.

\section{Conclusion}
\label{Conclusion}
In this paper, a SVM based dynamic optimization algorithm has been proposed. The basic idea is that knowledge can be grained from the process of solving the optimization problem, in other words we know which solutions belong to the POS in the past and which ones do not belong to the set. So such knowledge can be used to train a SVM classifier, and this classifier can be taken as a tool to classify the solutions for the dynamic optimization problem in the next time. Using this approach, a way is found to obtain an initial population, and this population can be fed into any population-based multi-objective optimization algorithm to solve the corresponding DMOPs. The experimental results show that the proposed approach is promising.

This research can be regarded as a starting point, and we will study how to train the SVM in an online mode and investigate it is possible to combine this method with transfer learning techniques \cite{jiang2017integration,8100935}. On the other hand, we will also study how to integrate this kind of method  with other approaches \cite{jiang2010embodied,jiang2014improving,jiang2012fuzzy} to solve the real world problems.


%

\section*{Acknowledgment}
This work was supported by the National Natural Science Foundation of China (No.61673328).

\ifCLASSOPTIONcaptionsoff
  \newpage
\fi



%

\bibliography{mybibtex}
%

\end{document}